\documentclass[sigconf]{acmart}
\usepackage{multirow} 
\usepackage{multicol}
\usepackage{algorithm}
\usepackage{algpseudocode}

\AtBeginDocument{%
  \providecommand\BibTeX{{%
    \normalfont B\kern-0.5em{\scshape i\kern-0.25em b}\kern-0.8em\TeX}}}

\copyrightyear{2023} 
\acmYear{2023} 
\setcopyright{acmlicensed}\acmConference[WWW '23]{Proceedings of the ACM Web Conference 2023}{May 1--5, 2023}{Austin, TX, USA}
\acmBooktitle{Proceedings of the ACM Web Conference 2023 (WWW '23), May 1--5, 2023, Austin, TX, USA}
\acmPrice{15.00}
\acmDOI{10.1145/3543507.3583337}
\acmISBN{978-1-4503-9416-1/23/04}

\begin{document}

\title{Semi-decentralized Federated Ego Graph Learning for Recommendation}

\author{Liang Qu}
\authornote{Both authors contributed equally to this research.}
\email{qul@mail.sustech.edu.cn}
\affiliation{%
  \institution{Southern University of Science and Technology}
  \city{Shenzhen}
  \country{China}
}

\author{Ningzhi Tang}
\authornotemark[1]
\email{11912521@mail.sustech.edu.cn}
\affiliation{%
  \institution{Southern University of Science and Technology}
  \city{Shenzhen}
  \country{China}
}

\author{Ruiqi Zheng}
\email{ruiqi.zheng@uq.net.au}
\affiliation{%
  \institution{The University of Queensland}
  \city{Brisbane}
  \country{Australia}
}

\author{Quoc Viet Hung Nguyen}
\email{henry.nguyen@griffith.edu.au}
\affiliation{%
  \institution{Griffith University}
  \city{Brisbane}
  \country{Australia}
}

\author{Zi Huang}
\email{huang@itee.uq.edu.au}
\affiliation{%
  \institution{The University of Queensland}
  \city{Brisbane}
  \country{Australia}
}

\author{Yuhui Shi}
\authornote{Corresponding Author}
\email{shiyh@sustech.edu.cn}
\affiliation{%
  \institution{Southern University of Science and Technology}
  \city{Shenzhen}
  \country{China}
}

\author{Hongzhi Yin}
\authornotemark[2]
\email{db.hongzhi@gmail.com}
\affiliation{%
  \institution{The University of Queensland}
  \city{Brisbane}
  \country{Australia}
}

\renewcommand{\shortauthors}{Liang and Ningzhi, et al.}

\begin{abstract}

Collaborative filtering (CF) based recommender systems are typically trained based on personal interaction data (e.g., clicks and purchases) that could be naturally represented as ego graphs. However, most existing recommendation methods collect these ego graphs from all users to compose a global graph to obtain high-order collaborative information between users and items, and these centralized CF recommendation methods inevitably lead to a high risk of user privacy leakage. Although recently proposed federated recommendation systems can mitigate the privacy problem, they either restrict the on-device local training to an isolated ego graph or rely on an additional third-party server to access other ego graphs resulting in a cumbersome pipeline, which is hard to work in practice.  In addition, existing federated recommendation systems require resource-limited devices to maintain the entire embedding tables resulting in high communication costs.

In light of this, we propose a semi-decentralized federated ego graph learning framework for on-device recommendations, named SemiDFEGL, which introduces new device-to-device collaborations to improve scalability and reduce communication costs and innovatively utilizes predicted interacted item nodes to connect isolated ego graphs to augment local subgraphs such that the high-order user-item collaborative information could be used in a privacy-preserving manner. Furthermore, the proposed framework is model-agnostic, meaning that it could be seamlessly integrated with existing graph neural network-based recommendation methods and privacy protection techniques.  To validate the effectiveness of the proposed SemiDFEGL, extensive experiments are conducted on three public datasets, and the results demonstrate the superiority of the proposed SemiDFEGL compared to other federated recommendation methods.

\end{abstract}

\ccsdesc[500]{Information systems~Recommender systems}

\keywords{federated ego graph learning, recommender systems, semi-decentralized learning, graph neural networks}

\maketitle

\section{Introduction}

\begin{figure}[t]
\centering
\includegraphics[width=0.5\textwidth]{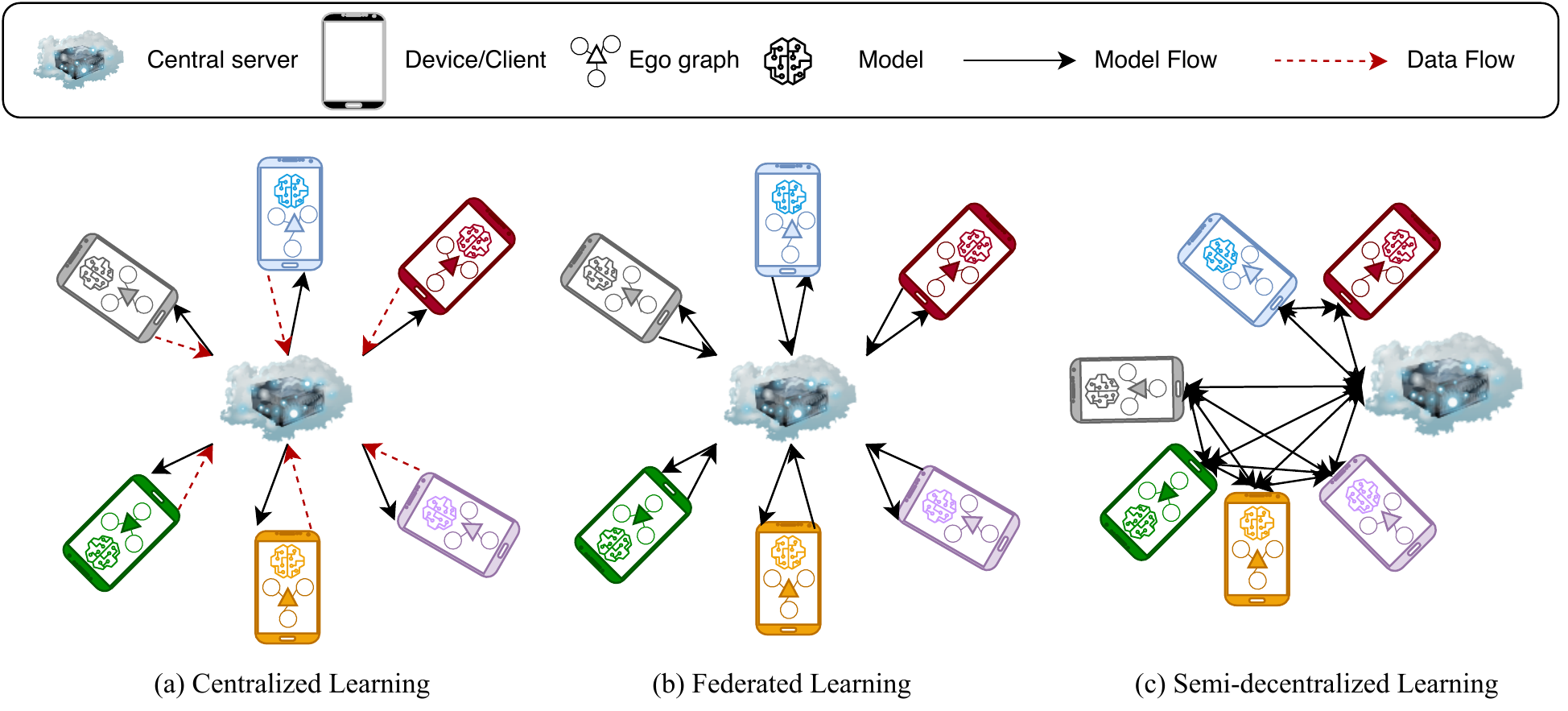} 
\caption{(a) Centralized learning (b) Federated learning (c) Semi-decentralized learning.}
\label{fig:overview}
\end{figure}

Recommender systems have been widely shown as an effective technique for helping users filter out the information they are not interested in.
The typical architectures of current recommender systems \cite{covington2016deep,guo2017deepfm,cheng2016wide} are that they collect all users' interaction data to train a powerful recommender model on a cloud server, as shown in Figure 1(a). However, with the increasing concerns for user privacy,  such centralized recommendation methods that hold personal data and models in the central server are most likely to break the new privacy regulations or laws such as General Data Protection Regulation (RDPG)\footnote{https://gdpr-info.eu/}. 


One natural solution to privacy issues is to keep users' data on their own devices and train models locally. However, a new problem that arises is that the local training data is very limited and thus insufficient to train an accurate model. Inspired by the effectiveness of federated learning \cite{li2020federated,yang2019federated} in solving privacy-preserving machine learning problems, federated recommender systems (FedRecs) \cite{yang2020federated,chen2018federated,ammad2019federated,flanagan2020federated,10.1145/3394486.3403176, wu2022federated,qi2020privacy,wang2021fast,chai2020secure,kim2016efficient} are proposed to collaboratively train recommendation models between local devices and a central server. Typically, as shown in Figure 1(b), FedRecs first train each local model on-device using the local dataset, and then sample a certain number of devices/clients to aggregate their local knowledge (e.g., model parameters/gradients) on the central server to learn a global model, which will then be redistributed back to each device for updating local models. 

Nevertheless, the current federated recommendation architectures still face the following challenges. (1) \textbf{Scalability:} The star-shaped federated recommender architecture \cite{ammad2019federated,chai2020secure,wang2021fast,10.1145/3394486.3403176} depends heavily on a central server to coordinate the device-server collaborations, making it infeasible to scale up to a large number of clients/users due to the communication bottleneck of the coordinator. 
There have been demonstrations with dozens and even hundreds of clients, but the complexity of a federated learning system comprising millions of mobile devices is simply inconceivable.
(2) \textbf{Constrained resources on client devices:}  Most existing FedRecs \cite{ammad2019federated,chai2020secure} require the client devices to maintain and store the whole item embedding table, which is hard to work in practice due to the constrained on-device resources. (3) \textbf{Cumbersome pipeline:} To achieve privacy protection or data augmentation, some FedRecs require an additional third-party server to implement encryption services \cite{kim2016efficient} or neighbor discovery \cite{wu2022federated},  resulting in a cumbersome pipeline and also increasing the risk of privacy leakage.

In light of the above challenges, we investigate the federated recommendation problem from a novel federated ego graph learning (FedEGL) perspective, in which the training data is treated as a set of user-centered ego graphs distributed over client devices, and each client is not allowed to share their ego graphs with either the cloud server or other clients.  In such a context, there is no global graph involved in the training process such that it is non-trivial to capture the high-order collaborative information. It is worth noting that federated graph learning under the recommendation scenario differs from the traditional federated graph learning (FedGL) \cite{he2021fedgraphnn,xie2021federated,zhang2021subgraph}. The major difference is that FedGL assumes that each device 
can have a large subgraph containing high-order graph information, while only first-order user-item interactions are available on each client device under the federated recommendation scenario. In addition, the subgraphs on different devices only have a few common nodes (i.e., anchor nodes),  while in the context of recommendations there are often a large number of common items between users/clients, and such collaborative information is critical to recommendation performance. Therefore, those FedGL methods cannot be applied to the federated recommendation scenario.

To address the dilemma of capturing the high-order graph information from isolated ego graphs, we propose a semi-decentralized federated ego graph learning framework (as shown in Figure 1(c)) for recommendation, named SemiDFEGL, which introduces device-to-device collaborations to improve scalability as well as reduce communication costs, and also innovatively utilizes predicted interacted item nodes (called fake common items) to connect isolated ego graphs to construct local high-order subgraphs.  Specifically, a client first learns an embedding representation for the local ego graph and then uploads the ego graph embedding to the central server. Based on the received ego graph embeddings and the item embedding table,  the central server clusters users/items into a number of groups. To exploit the high-order graph information, items assigned to each group are called predicted interacted item nodes (also called fake common items) and are utilized to connect to each user within the group thus forming a local collaboration/communication graph on which clients/users  can directly communicate with their neighbors to perform collaborative learning, making this semi-decentralized federated architecture more scalable. To be more specific, each client/user only needs to share aggregated node embeddings with other clients having common fake item nodes (i.e., neighbor clients) by the  message-passing mechanism. In this way, users/clients do not need to share their own ego graphs. After finishing local collaborative training, each user/client only needs to upload the updated item embeddings (including embeddings of both positive and selected negative items) to the central server for updating the global item embedding table. Note that each client does not need to download and store the global item embedding table on their resource-constrained devices as the final recommendations are produced on the central server by calculating the relevance between the ego graph embedding and each item embedding.   Furthermore, the proposed framework can accommodate almost all existing  graph neural networks (GNNs) based recommenders and seamlessly integrate with existing privacy protection techniques (e.g., local differential privacy \cite{joseph2018local}).

The contributions of this paper are summarized as follows:
\begin{itemize}
    \item We propose a semi-decentralized federated ego graph learning framework for recommendation, named SemiDFEGL, which introduces  a novel device-to-device collaborative learning mechanism to improve scalability and reduce communication costs of the conventional FedRecs. The proposed framework is model-agnostic, which can accommodate almost all existing  graph neural networks (GNNs) based recommenders and seamlessly integrate with existing privacy protection techniques such as LDP.
    \item We propose a novel local ego graph augmentation method to leverage predicted interacted items to connect isolated ego graphs so that the high-order collaborative information could be exploited in a privacy-preserving manner. 
    \item Extensive experiments are conducted on three public datasets, and the experimental results demonstrate that the proposed framework could outperform other federated recommendation methods.
\end{itemize}

The rest of this paper is organized as follows. Section 2 will review the main related works, and Section 3 will detail the proposed framework. The experiments are introduced in Section 4, followed by a conclusion in Section 5.

\section{Related Work}

\subsection{Recommender Systems}
Recommender systems have been shown as an effective technique for helping users filter out the information they are not interested in based on their feature information and historical interaction data (e.g., ratings and clicks). Early approaches mainly include collaborative filtering \cite{zhao2010user,sarwar2001item}, matrix factorization \cite{koren2009matrix,ramlatchan2018survey} and factorization machines \cite{rendle2010factorization,juan2016field}. Recently, deep learning \cite{lecun2015deep} has achieved significant progress in computer vision \cite{krizhevsky2012imagenet} and natural language processing \cite{hinton2012deep} due to the capabilities of capturing high-order non-linear features. Thus, there are a lot of works to apply it to recommender systems, such as deep neural network-based recommendation \cite{covington2016deep,guo2017deepfm,yin2015joint,cheng2016wide,10.1145/3579355,10.1145/3477495.3532060}, graph neural network-based recommendation \cite{ying2018graph,wu2020graph}, and attention-based recommendation \cite{kang2018self,zhou2018deep}. However, as mentioned above, these methods are typically deployed in a centralized manner and collect all user data to train the model, which could suffer from data privacy, resource consumption, and response latency issues.

\begin{figure*}[t]
\centering
\includegraphics[width=0.6\textwidth]{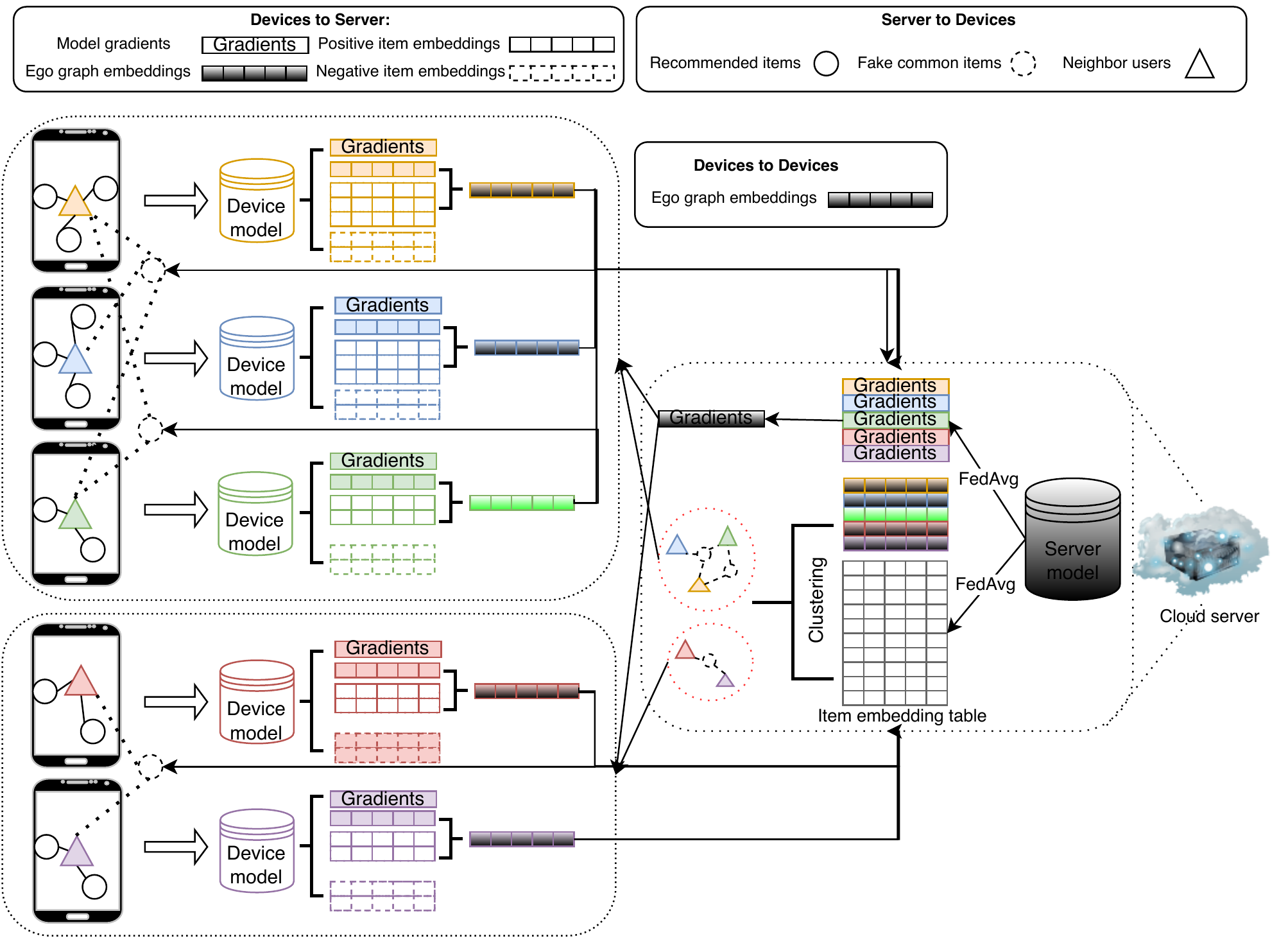} 
\caption{The overview of SemiDFEGL. The user-item interaction data, i.e., the ego graph, and item data are stored on the device side and cloud side, respectively. The device learns the embedding representation for the local ego graph and uploads it to the central server. The central server clusters users/items into a number of groups based on their embeddings. Items assigned to each group are called predicted interacted item nodes and are utilized to connect to each user within the group to form a higher-order local subgraph.}
\label{fig:overview}
\end{figure*}

\subsection{Federated Recommender Systems}
Inspired by the effectiveness of federated learning in addressing privacy-preserving machine learning problems, federated recommender systems \cite{zhang2021graph,wang2021fast,zhang2022pipattack,imran2022refrs,wang2020next} are proposed to train on-device models in a device-to-server collaboration manner.
For example, 
MetaMF \cite{lin2020meta} proposes a meta network to generate private item embeddings and rating prediction models deployed on user devices, and each device uploads the gradient information to the server for updating the meta network.
FCF \cite{ammad2019federated} extends the collaborative filter (CF) to the federated model, which achieves comparable performance against standard CF while fully preserving the user's privacy. 
FedFast \cite{10.1145/3394486.3403176} introduces the active sampling and aggregation strategies to accelerate the convergence of local models.
One similar work to ours is FedPerGNN \cite{wu2022federated} which also explores to learn privacy-preserving recommendations from a federated graph learning perspective. However, one major difference from our approach is that FedPerGNN requires a trusted third party to find users who have co-interacted items, which results in a cumbersome pipeline, and it is difficult to find such a trusted third-party server in practice. There are also some works allowing devices to collaboratively learn with its neighbor devices in a fully device-to-device (D2D) fashion, namely decentralized recommender systems \cite{long2022decentralized,beierle2019collaborating,beierle2020mobrec,chen2018privacy}. However, such methods only rely on D2D collaborations and generally require that each device stores and maintains huge chunks of non-privacy item data, which is unnecessary and hard to work in practice due to the constrained resources.

\section{Proposed Method}

\subsection{Preliminary}
\begin{itemize}
    \item \textbf{User-centered ego graph:} An user-centered ego graph $\mathcal{G}=\{u,V,E\}$ consists of a user node $u$, a set of items $V=\{v_{i}\}_{i=1}^{n}$, and a set of edges/interactions $E=\{e_{j}\}_{j=1}^{n}$, where $v_{i}$ and $e_{j}$ represent a item node and an interaction (e.g., clicks and ratings) between user $u$ and item $v_{j}$, respectively. $n$ is the number of items that user $u$ has interacted with on a local device. 
    \item \textbf{Federated ego graph learning (FedEGL) for recommendation:} Suppose that there are a set of devices $\mathcal{D}=\{D_{k}\}_{k=1}^{K}$, where $K$ is the number of devices, and each device $D_{k}=\{\mathcal{G}_{k},f(\Theta_{k})\}$ consists of an ego graph $\mathcal{G}_{k}$ and an on-device model $f(\Theta_{k})$ parameterized by $\Theta_{k}$. FedEGL aims to predict the user's preferences based on the user-item interaction data $\mathcal{G}_{k}$ in a privacy-preserving manner. It is worth noting that each device is not allowed to share its ego graph with either the central server or other devices.
\end{itemize}

\subsection{SemiDFEGL}
To address the dilemma of exploiting the high-order graph information from isolated ego graphs, and improve the scalability as well as reduce communication costs of current federated recommendation architectures, we propose a semi-decentralized federated ego graph learning framework for recommendation, named SemiDFEGL, which is illustrated in Figure \ref{fig:overview}. It consists of two main collaborative learning strategies: the centralized device-to-server (D2S) collaboration and the decentralized device-to-device (D2D) collaboration. Concretely, each client learns a local ego graph embedding and then uploads it to the central server. The central server clusters users/items into a number of groups based on their embeddings. Items assigned to each group are called predicted interacted item nodes (also called fake common items) and are utilized to connect to each user within the group to form a higher-order local subgraph. In this way, clients within the same group can directly communicate with their neighbors to perform D2D collaborations by sharing aggregated node embeddings with other clients rather than their ego graphs via common fake items. After that, clients only need to upload item embeddings (including embeddings of both positive and selected negative items) to the central server for updating the global item embedding table. 


\subsubsection{Local Ego Graph Augmentation} For the traditional GNN-based recommendation models \cite{he2020lightgcn} that maintains a global user-item interaction graph on the central server, the model can directly learn user/item embeddings from high-order user-item collaborative information over the global graph. However, it will inevitably lead to a high risk of user privacy leakage. In order to protect the privacy of the user $u$, one natural solution is to keep personal interaction data that is represented as the ego graph $\mathcal{G}_{u}$ on its own client. Thus, one major technical challenge is how to exploit high-order user-item collaborative information without actual data sharing. To address this problem, we propose to utilize a central server to generate a set of fake common items to connect users thus forming a local collaboration subgraph. 

Specifically, each client first learns an embedding representation for the local ego graph by the message-passing mechanism of GNN as follows:

\begin{equation}
    \mathbf{e}_{u}^{(k+1)} = AggFun(\mathbf{e}_{u}^{(k)},\{\mathbf{e}^{(k)}_{i}\}_{i \in \mathcal{N}_{(u)}})
\end{equation}

where $\mathbf{e}_{u}$ and $\mathbf{e}_{i}$ are user and item embeddings, respectively. $k$ is the number of layers, and $\mathcal{N}_(u)$ represents the set of neighbor nodes of user $u$. $AggFun(\cdot)$ is the aggregation function which could be defined by different strategies of GNN variants such as graph convolution network (GCN \cite{kipf2016semi}), graph attention networks (GAT \cite{velivckovic2017graph}), and Light Graph Convolution (LGC \cite{he2020lightgcn}). Note that the proposed framework is model-agnostic, meaning that it could be integrated with most existing GNN variants. In this section, we choose LGC as an example to introduce our method due to its superiority in the recommendation problem. Thus, the user/item embeddings could be learned as follows \cite{he2020lightgcn}:
\begin{equation}
\begin{aligned}
&\mathbf{e}_u^{(k+1)}=\sum_{i \in \mathcal{N}_{(u)}} \frac{1}{\sqrt{\left|\mathcal{N}_{(u)}\right|} \sqrt{\left|\mathcal{N}_{(i)}\right|}} \mathbf{e}_i^{(k)} \\
&\mathbf{e}_i^{(k+1)}=\sum_{u \in \mathcal{N}_{(i)}} \frac{1}{\sqrt{\left|\mathcal{N}_{(i)}\right|} \sqrt{\left|\mathcal{N}_{(u)}\right|}} \mathbf{e}_u^{(k)}
\end{aligned}
\label{equ:ego}
\end{equation}
For each ego graph, as it only has first-order user-item interaction information, we can directly obtain its user embedding without layer combination. Since there is no ego graph augmentation in the initialization phase, we directly replace the ego graph embedding with the user's embedding, i.e., $\mathbf{e}_{ego} = \mathbf{e}_{u}$ and upload it to the central server.

Now, the central server has obtained the uploaded ego graph embeddings which represent users' historical preferences. In order to connect isolated ego graphs, we propose to employ a co-clustering method to cluster users and items into a number of groups based on ego graph embeddings $\mathcal{E}_{U}=\{\mathbf{e_{u}}\}_{u=1}^{N}$ and item embedding table $\mathcal{E}_{V}=\{\mathbf{e_{v}}\}_{v=1}^{M}$, where $N$ and $M$ represent the number of users and the number of items, respectively. In particular, considering the characteristics of recommendation systems, i.e., there may be an item that belongs to more than one group at the same time. We utilize the fuzzy c-means \cite{lkeski2003towards,xu2012exploration} to learn the membership $P_{ij}$ of entry $\mathbf{x}_{i}$ (an ego graph embedding or an item embedding) in the group $j$ by optimizing the following criterion function \cite{xu2012exploration} in central server:

\begin{equation}
J_m(P, C)=\sum_{i=1}^{M+N} \sum_{j=1}^C\left(P_{i j}\right)^l d\left(\mathbf{x}_i, \mathbf{c}_j\right)^2
\label{equ:clustering}
\end{equation}

where $\mathbf{c}_{j}$ is the center of group $j$. $d(\cdot)$ is the distance function, and we employ Euclidean distance in this paper. $l$ is the weighting exponent to control the fuzziness of the resulting partition. During each iteration in a central server, $P$ and $C$ are updated as follows until the termination condition is met.
\begin{equation}
P_{i j}=\left(d\left(\mathbf{x}_i, \mathbf{c}_j\right)\right)^{2 /(1-l)} /\left[\sum_{k=1}^c\left(d\left(\mathbf{x}_i, \mathbf{c}_k\right)\right)^{2 /(1-l)}\right]
\end{equation}
\begin{equation}
\mathbf{c}_j=\left[\sum_{i=1}^{M+N} P_{i j}^l \mathbf{x}_i\right] /\left[\sum_{i=1}^{M+N} P_{i j}^l\right]
\end{equation}

After that, for each user's membership, we select the element with the largest value to ensure that each user/client belongs to only one group. For the membership of each item, we pick the top-k largest elements so that the same item can belong to more than one group. Thus, the items assigned to each group are called predicted interacted item nodes (also called fake common items) and are utilized to connect to each user within the group to form a higher-order local subgraph.

\begin{figure}[t]
\centering
\includegraphics[width=0.5\textwidth]{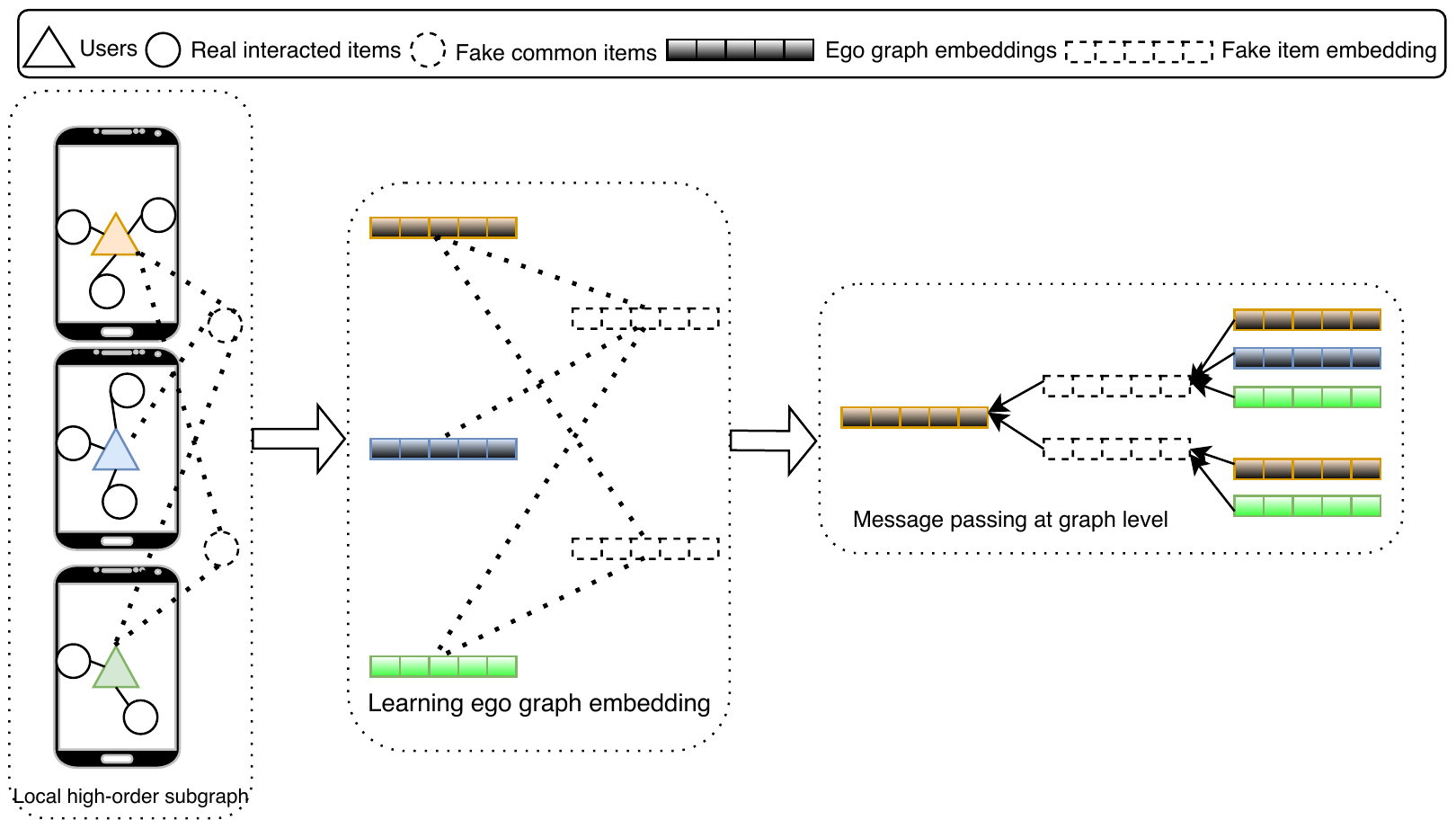} 
\caption{The illustration of device-to-device collaborations via message passing at graph level.}
\label{fig:d2d}
\end{figure}

\subsubsection{Device-to-Device Collaborations}
Although we have constructed higher-order subgraphs in each group, each device is not allowed to share its own ego graph with its neighboring clients due to privacy concerns. The information shared by the clients in the group is that each client only knows the common fake items and the clients connected to them. Therefore, we propose to perform device-to-device message passing at the graph level rather than at the node level. In particular, as shown in Figure \ref{fig:d2d}, each device first learns its own ego graph embedding based on equation \ref{equ:ego}, and shares it with other neighbor clients via fake common items. In this way, the message-passing mechanism of GNN can be performed in a standard way, 
such as the layer combination operation of LightGCN \cite{he2020lightgcn} as follows:
\begin{equation}
\mathbf{e}_u=\sum_{k=0}^K \alpha_k \mathbf{e}_u^{(k)} ; \quad \mathbf{e}_i=\sum_{k=0}^K \alpha_k \mathbf{e}_i^{(k)}
\label{equ:d2d}
\end{equation}
where  $\alpha_k$ is the hyperparameter representing the importance of the $k$-th layer. Thus, each client can learn the final user embedding $\mathbf{e}_{u}$ and the item embedding $\mathbf{e}_{i}$ locally that contains the high-order graph information.

\subsubsection{Device-to-Server Collaborations}
After learning node embedding locally, we perform the device-to-server collaborations by sampling devices from each group and uploading item embeddings to the central server for updating the global item embedding table. In particular, to protect the user's interaction data, each client samples a number of items that it has not interacted with and randomly generates their embeddings $\mathbf{e}_{i}^{-}$ with the same mean and variance as the real interacted item embedding $\mathbf{e}_{i}^{+}$. Then, the embeddings of both positive and negative items are uploaded to the central server to perform the FedAvg \cite{mcmahan2017communication}. Furthermore, since the embeddings could also leak the user's privacy, we clip the uploaded embeddings based on L1-norm with a threshold $\delta$, and integrate LDP with embeddings to further protect user privacy as follows:
\begin{equation}
\hat{\mathbf{e}}_i=\operatorname{clip}\left(\hat{\mathbf{e}}_i, \delta\right)+\operatorname{Laplace}(0, \lambda)
\label{equ:ldp}
\end{equation}
where $\hat{\mathbf{e}}_i = \{\mathbf{e}_{i}^{+},\mathbf{e}_{i}^{-}\}$ are the embeddings that need to be uploaded to the central server, and $\lambda$ is the noise scale.

\subsubsection{Model Prediction and Training}
The model prediction, i.e., the personalized recommendation, is produced on the central server by calculating the ranking score $y_{ui}$ of the upload ego graph embedding of the user and each item embedding as follows:
\begin{equation}
    y_{ui} = \mathbf{e}_{ego}^{T}\mathbf{e}_{i}
\end{equation}
In this way, each client does not need to download and store the global item embedding table on their resource-constrained devices.

For the model training on each client, we utilize the Bayesian Personalized Ranking (BPR) \cite{rendle2012bpr} as the loss function $\mathcal{L}$ as follows:
\begin{equation}
\mathcal{L}=- \sum_{i \in \mathcal{N}_u} \sum_{j \notin \mathcal{N}_u} \ln \sigma\left(y_{u i}-y_{uj}\right)+\lambda\left\|\mathbf{e}\right\|^2
\end{equation}
where $\lambda$ is a hyperparameter that controls the strength of the $L_{2}$ regularization, and $\mathbf{e}=\{\mathbf{e}_{i},\mathbf{e}_{u}\}$ is the trainable user/item embeddings. Note that since each client knows the items it has actually interacted with itself, the fake common items can be considered as negative samples in the model training process.


\section{Experiments}

To validate the effectiveness of the proposed method, we conduct extensive experiments to answer the following research questions (RQs):
\begin{itemize}
    \item \textbf{RQ1}: How does the proposed method perform compared with other federated recommendation methods?
    \item \textbf{RQ2:} Can the proposed method reduce communication costs?
    \item \textbf{RQ3:} How do different components affect the performance of the proposed method?
    \item \textbf{RQ4:} How do different hyperparameters influence the performance of the proposed method?
    \item \textbf{RQ5:} Can the proposed method be integrated with other GNN-based recommendation methods?
\end{itemize}

\begin{table}[htbp]
 
  \centering
  \caption{The statistics of datasets.}
    \begin{tabular}{c|c|c|c}
    \toprule
    Dataset & MovieLens-1M & Yelp2018 & Gowalla \\
    \midrule
    \midrule
    \#User & 6,040 & 31,668 & 29,858 \\
    \#Item & 3,900 & 38,048 & 40,981 \\
    \#Interactions & 1,000,209 & 1,561,406 & 1,027,370 \\
    \bottomrule
    \end{tabular}%
  \label{tab:dataset}%
\end{table}%

\begin{table*}[htbp]
  \centering
  \caption{The top-K recommendation performance of SemiDFEGL and baselines on three datasets.}
    \begin{tabular}{c|c|cc|cc|cc}
    \toprule
    \multicolumn{2}{c|}{\multirow{2}[4]{*}{Method}} & \multicolumn{2}{c|}{MovieLen-1M} & \multicolumn{2}{c|}{Yelp2018} & \multicolumn{2}{c}{Gowalla} \\
\cmidrule{3-8}    \multicolumn{2}{c|}{} & Recall@20 & NDCG@20 & Recall@20 & NDCG@20 & Recall@20 & NDCG@20 \\
    \midrule
    \midrule
    \multirow{2}[2]{*}{Cloud-based Recs} & NeuMF & 0.2402 & 0.3641 & 0.0572 & 0.0489 & 0.1506 & 0.1307 \\
          & LightGCN & 0.2559 & 0.3859 & 0.0626 & 0.0511 & 0.18  & 0.1523 \\
    \midrule
    \midrule
    \multirow{4}[2]{*}{FedRecs} & FCF   & 0.2301 & 0.3556 & 0.0517 & 0.0417 & 0.1434 & 0.1211 \\
          & FedMF & 0.2397 & 0.3632 & 0.0491 & 0.0402 & 0.1432 & 0.1227 \\
          & MetaMF & 0.2135 & 0.3342 & 0.0346 & 0.0283 & 0.1005 & 0.083 \\
          & FedPerGNN & 0.2435 & 0.3686 & 0.0449 & 0.0445 & 0.1446 & 0.1127 \\
    \midrule
    \midrule
    Ours  & SemiDFEGL & 0.2486 & 0.3771 & 0.0583 & 0.0491 & 0.168 & 0.1372 \\
    \bottomrule
    \end{tabular}%
  \label{tab:topk}%
\end{table*}%

\subsection{Experimental Settings}

\subsubsection{Datasets}
We validate baselines and the proposed method on three widely-used public datasets, i.e., MovieLens-1M\footnote{https://grouplens.org/datasets/movielens/1m/}, Yelp2018\footnote{https://www.yelp.com/dataset/challenge} and Gowalla \cite{liang2016modeling}. The statistics of the datasets are summarized in Table 1. All these three datasets are composed of historical user-item interactions. Following \cite{10.1145/3038912.3052569,KGAT19}, we filter out those users and items with less than 20/10/10 interactions for MovieLens-1M/Yelp2018/Gowalla, respectively. All three datasets are randomly split into training sets, validation sets, and test sets as the ratio 8:1:1, respectively.

\subsubsection{Baselines}
We compare SemiDFEGL with six baselines from two categories including the cloud-based recommendation methods and federated recommendation methods.

\begin{itemize}
    \item \textbf{Cloud-based recommendation methods:}
    \begin{itemize}
        \item \textbf{NeuMF} \cite{10.1145/3038912.3052569}: It is a representative deep recommendation method that uses DNN to replace the dot product operation to capture non-linearity in the implicit feedback.
        \item \textbf{LightGCN} \cite{he2020lightgcn}: It utilizes graph neural networks to learn the user/item embeddings via the linear neighborhood aggregation mechanism on the user-item interaction graph, which is the state-of-the-art recommendation method.
    \end{itemize}
    \item \textbf{Federated recommendation methods:}
    \begin{itemize}
        \item \textbf{FCF} \cite{ammad2019federated}: It extends CF to the federated model, which achieves comparable performance against standard CF while fully preserving the user's privacy.
        \item \textbf{FedMF} \cite{chai2020secure}: It proposes a user-level distributed matrix factorization framework to well protect users' privacy by the homomorphic encryption technique, which only requires users to upload gradients to the cloud.
        \item \textbf{MetaMF} \cite{lin2020meta}: It is a federated learning-based matrix factorization method that has a similar motivation to us, i.e., generating personalized models for each device. It learns a meta network on the cloud server, and the meta network utilizes the collaborative vector to generate private and personalized item embeddings and prediction models for each user.
        \item \textbf{FedPerGNN} \cite{wu2022federated}: It is a GNN-based federated recommendation method, which proposes to use a trusted third-party server to assign neighbors who have co-interacted items to each user such that the high-order graph information could be captured.
    \end{itemize}
\end{itemize}

\subsubsection{Evaluation metrics}
We use two widely used evaluation metrics \textit{Recall@20} and \textit{NDCG@20} (Normalized Discounted Cumulative Gain) to measure the recommendation performance\cite{yao2021device,he2020lightgcn,10.1145/3038912.3052569}. Following \cite{he2020lightgcn}, we treat all items that have not interacted with a user as candidates, and the average results over all users are reported.

\subsubsection{Hyper-parameter Setting}
The hyperparameters of the proposed SemiDFEGL are set as follows. The user/item embeddings are initialized with the Xavier method \cite{glorot2010understanding}, and the dimension is 64 for all methods. We utilize Adam optimizer \cite{kingma2014adam} with an initial learning rate of 0.001, and the weight decay is 0.0001. The number of fake common item nodes and the number of layers are 1 and 4 for all datasets, respectively. The number of groups is 100 on MovieLens-1M and 200 on the other two datasets. The baselines are implemented by the codes provided by the authors. 

\begin{figure*}[t]
\centering
\includegraphics[width=0.7\textwidth]{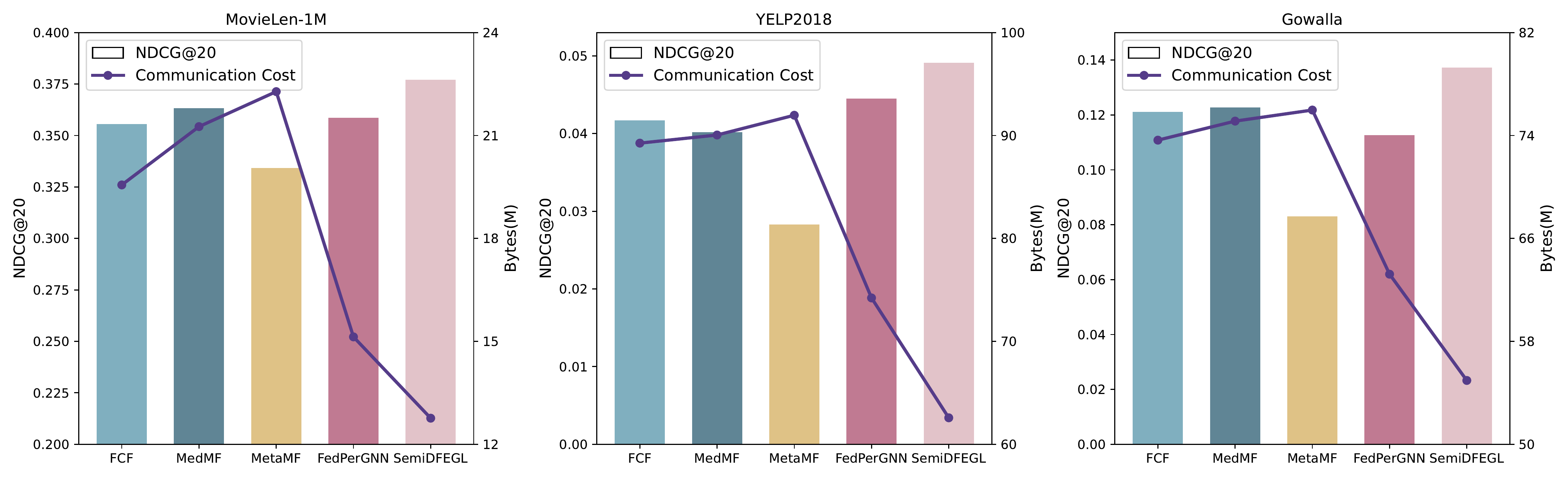} 
\caption{The model performance and communication costs for different federated recommendation methods.}
\label{fig:cc}
\end{figure*}

\begin{table*}[htbp]
  \centering
  \caption{The ablation study results with respect to fake common item nodes.}
    \begin{tabular}{c|cc|cc|cc}
    \toprule
    \multirow{2}[3]{*}{Method} & \multicolumn{2}{c|}{MovieLen-1M} & \multicolumn{2}{c|}{Yelp2018} & \multicolumn{2}{c}{Gowalla} \\
\cmidrule{2-7}          & Recall@20 & NDCG@20 & Recall@20 & NDCG@20 & Recall@20 & NDCG@20 \\
    \midrule
    \midrule
    SemiDFEGL w/o fake items & 0.2314 & 0.3502 & 0.0467 & 0.04  & 0.149 & 0.1277 \\
    SemiDFEGL & \textbf{0.2486} & \textbf{0.3771} & \textbf{0.0583} & \textbf{0.0491} & \textbf{0.168} & \textbf{0.1372} \\
    \bottomrule
    \end{tabular}%
  \label{tab:ablation}%
\end{table*}%

\subsection{Top-k recommendation (RQ1)}
We first validate the effectiveness of the proposed SemiDFEGL on the top-k recommendation task. The experimental results are shown in Table \ref{tab:topk}. We can observe that:
\begin{itemize}
    \item The cloud-based recommendation methods achieve better performance than the federated recommendation methods in most cases. A major reason is that FedRecs methods need encryption services to protect user privacy, which introduces additional noise, resulting in limiting the model capability on learning user/item representations and model optimization.
    \item The graph-based methods (i.e., LightGCN, FedPerGNN, and SemiDFEGL) achieve better performance than MF-based methods (i.e., NeuMF, FedMF, and MetaMF). The possible reason is that GNN could better capture the implicit collaborative information between users and items by the  message-passing mechanism, which further validates the reasonability that we model the federated recommendation problem from the federated ego graph learning perspective.
    \item SemiDFEGL significantly outperform other federated recommendation methods on three datasets and achieve competitive performance against cloud-based methods. For example, it exceeds FedPerGNN by $2.09\%$ and $2.3\%$ with respect to \textit{Recall@20} and \textit{NDCG20} respectively on MovieLens-1M. We attribute it to the fact that SemiDFEGL can well learn high-order user-item collaborative information via device-to-device collaborations.
\end{itemize}

\subsection{Communication costs (RQ2)}
Recall that we argue that the communication bottleneck of the central server makes the current federated recommendation framework infeasible to scale up to a large number of clients. Thus, we compare the model performance with their communication costs in this section. Specifically, we use the number of parameters to be uploaded and downloaded in each training round to measure the communication cost of the model. The experiments are conducted on three datasets, and the results are shown in Figure \ref{fig:cc}, from which we can observe that:
\begin{itemize}
    \item Our model requires minimal communication costs but achieves the best performance. We attribute this to the fact that we do not need each device to maintain the global item embedding table as the final recommendations are produced on the central server.
    \item MF-based federated recommendation methods (i.e., FedMF, and MetaMF) require larger communication costs than GNN-based federated recommendation methods (i.e., FedPerGNN, and SemiDFEGL). It is reasonable because the former requires that the user client needs to upload the entire embedding table in each training round for protecting user interaction data. 
\end{itemize}

\subsection{Ablation Study (RQ3)}
In this section, we target to demonstrate the effect of the fake common items. In particular, we implement the SemiDFEGL without fake common items, denoted as \textit{SemiDFEGL w/o fake items}. In this way, each client can only use its ego graph to train the local model, and device-to-device collaborations are also not available. The experiments are conducted on three datasets, and other settings are the same as the above experiments. The results are shown in Table \ref{tab:ablation}, from which we can observe that:

\begin{itemize}
    \item When the model has no fake common items, the performance of the model decreases substantially. This is reasonable because each isolated subgraph can only learn its first-order graph information and there is no collaboration between subgraphs, which further validates the necessity of our proposed ego graph augmentation strategy.
\end{itemize}

\begin{table*}[htbp]
  \centering
  \caption{Performance of SemiDFEGL with different GNN variants.}
    \begin{tabular}{c|c|c|cc|cc}
    \toprule
    \multirow{2}[4]{*}{Methods} & \multicolumn{2}{c|}{MovieLens-1M} & \multicolumn{2}{c|}{Yelp2018} & \multicolumn{2}{c}{Gowalla} \\
\cmidrule{2-7}          & \multicolumn{1}{l}{Recall@20} & \multicolumn{1}{l|}{NDCG@20} & \multicolumn{1}{l}{Recall@20} & \multicolumn{1}{l|}{NDCG@20} & \multicolumn{1}{l}{Recall@20} & \multicolumn{1}{l}{NDCG@20} \\
    \midrule
    \midrule
    GCN   & 0.2466 & 0.03764 & 0.0574 & 0.048 & 0.1663 & 0.1355 \\
    GAT   & 0.2472 & 0.3765 & 0.0569 & 0.0474 & 0.1672 & 0.1362 \\
    GraphSAGE & 0.2467 & 0.3764 & 0.0571 & 0.0475 & 0.167 & 0.1358 \\
    SemiDFEGL (LightGCN) & \textbf{0.2486} & \textbf{0.3771} & \textbf{0.0583} & \textbf{0.0491} & \textbf{0.168} & \textbf{0.1372} \\
    \bottomrule
    \end{tabular}%
  \label{tab:model-agnostic}%
\end{table*}%

\subsection{Hyper-parameter Sensitivity Analysis (RQ4)}
\begin{figure}[t]
\centering
\includegraphics[width=0.4\textwidth]{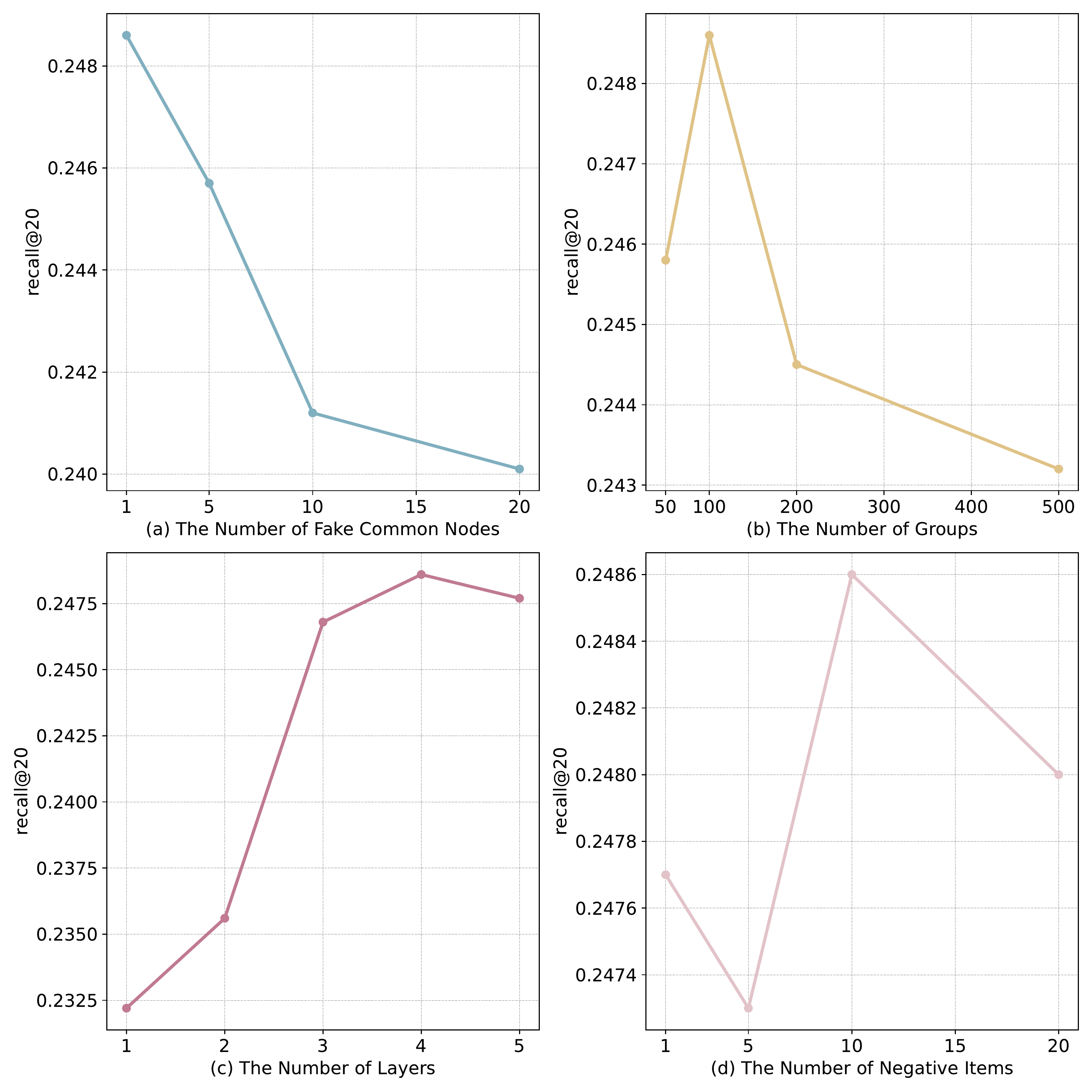} 
\caption{The performance of model with different hyper-parameter settings. (a) The number of fake common nodes. (b) The number of groups. (c) The number of layers. (d) The number of negative items.}
\label{fig:hyper}
\end{figure}

To explore the effect of four crucial hyperparameters, i.e., the number of fake common items $\#Fake nodes=\{1,5,10,20\}$, the number of groups $\#Groups=\{50,100,200,500\}$, the number of layers $\#Layers=\{1,2,3,4,5\}$, and the number of negative items $\#NegativeItems=\{1,5,10,20\}$, we conduct experiments on MovieLens-1M dataset, and report the model performance with different hyper-parameter changes on Figure \ref{fig:hyper}. We can observe that:
\begin{itemize}
    \item As the number of fake common items increases, the performance of the model decreases. This is reasonable because the fake common items could connect isolated subgraphs for exploiting higher-order graph information. However, fake common items can somewhat disrupt the original data distribution and introduce additional noise.
    \item The model achieves the best performance when the number of groups is 100. The possible reason for this is that when the number of groups is small, the number of fake common items within each group is high, which brings in more noise and thus reduces the model performance. On the other hand, when the number of groups is large, the number of users in each cluster is small, resulting in insufficient learning of higher-order graph information.
    \item The model achieves the best performance when the number of layers of the network is 4. This is explained by that when the number of layers is low, the model is unable to learn the higher-order structural information of the graph. On the other hand, when the number of layers is high, the over-smoothing problem \cite{li2018deeper} of GNNs may be introduced.
    \item Although the model can achieve better results when the number of negative items is small, the privacy of users may be at risk of leakage. With a slight increase in the number of negative items, additional noise is introduced which reduces the performance of the model. However, when the number of negative items is large, the negative effects may be offset.
\end{itemize}

\subsection{Model-agnostic study (RQ5)}
This section aims to explore if the proposed semi-decentralized federated ego graph learning framework can accommodate other GNN methods. To this end, we implement the proposed method with various GNN variants including graph convolution network (GCN \cite{kipf2016semi}), graph attention networks GAT \cite{velivckovic2017graph}), and GraphSAGE \cite{hamilton2017inductive}. Experiments are conducted on MovieLen-1M dataset, and the one-hot ID vector is utilized as the node attributes. Experimental results are shown in Table \ref{tab:model-agnostic}. From the results, we can observe that:
\begin{itemize}
    \item The model integrated with LightGCN achieves the best performance among the four variants. The possible reason is that the other three variants use the nonlinear layer to learn the embeddings of user and item nodes, but the nodes of the user-item bipartite graph have only one-hot ID features, leading to increased difficulty in optimizing the nonlinear layer.
    \item The attention-based methods (i.e., GAT and LightGCN) can achieve better performance than the other two variants. The performance improvement could be attributed to the attention mechanism that can well capture the importance of different neighbor nodes.
\end{itemize}

\section{Conclusion}
In this work, we investigated the federated recommendation problem from a novel federated ego graph learning perspective, and proposed a novel semi-decentralized federated ego graph learning framework for recommendation, named SemiDFEGL, which introduces device-to-device collaborations to improve scalability as well as reduce communication costs. To address the dilemma of capturing the high-order graph information from isolated ego graphs, we innovatively utilized fake common items to connect isolated ego graphs to construct local high-order subgraphs. Furthermore, the proposed framework can accommodate almost all existing GNN-based recommenders and seamlessly integrate with existing privacy protection techniques. To validate the effectiveness of the proposed method, comprehensive experiments have been conducted on three public datasets, and empirical results have demonstrated the superiority of the proposed method against federated recommendation methods. For the future work, we plan to explore extending our framework to non-graph-based recommendation models.


\section{Acknowledgement}
This work is supported by Australian Research Council Future Fellowship (Grant No.FT210100624), Discovery Project (Grant No. DP190101985), Discovery Early Career Research Award (Grant No. DE200101465), the National Natural Science Foundation of China (Grant No. 61761136008), the Shenzhen Fundamental Research Program (Grant No. JCYJ20200109141235597), the Shenzhen Peacock Plan  (Grant No. KQTD2016112514355531), the Guangdong Basic and Applied Basic Research Foundation (Grant No. 2021A1515110024), and the Program for Guangdong Introducing Innovative and Entrepreneurial Teams (Grant No. 2017ZT07X386).

\bibliographystyle{ACM-Reference-Format}
\bibliography{main}

\appendix

\end{document}